\documentclass{article}
\usepackage[a4paper, total={6in, 8in}]{geometry}
\usepackage{graphicx} 
\usepackage{amsmath}
\usepackage{amsfonts}
\usepackage{bm}
\usepackage{cleveref}
\usepackage{xcolor}
\usepackage{bibentry}
\usepackage{bbm}
\usepackage[numbers]{natbib} 
\usepackage{rotating}

\usepackage{todonotes}

\newcommand{\ignore}[1]{}
\newcommand{\nobibentry}[1]{{\let\nocite\ignore\bibentry{#1}}}

\newcommand{\mand}{\; \; {\rm and} \; \;}

\newcommand{\vbar}{\,|\, }
\providecommand{\keywords}[1]
{
  \small	
  \textbf{\textit{Keywords:}} #1
}

\newif\ifexclude

\title{Fast Gibbs sampling for the local and global trend Bayesian exponential smoothing model}

\author{
  Xueying Long\\
  \texttt{Xueying.Long1@monash.edu}
  \and
  Daniel F. Schmidt\\
  \texttt{Daniel.Schmidt@monash.edu}
  \and
  Christoph Bergmeir\\
  \texttt{bergmeir@ugr.es}
  \and
  Slawek Smyl\\
  \texttt{slawek.smyl@gmail.com}
}

\begin{document}

\maketitle

\begin{abstract}
   In Smyl et al. [Local and global trend Bayesian exponential smoothing models. International Journal of Forecasting, 2024.], a generalised exponential smoothing model was proposed that is able to capture strong trends and volatility in time series. This method achieved state-of-the-art performance in many forecasting tasks, but its fitting procedure, which is based on the NUTS sampler, is very computationally expensive. In this work, we propose several modifications to the original model, as well as a bespoke Gibbs sampler for posterior exploration; these changes improve sampling time by an order of magnitude, thus rendering the model much more practically relevant. The new model, and sampler, are evaluated on the M3 dataset and are shown to be competitive, or superior, in terms of accuracy to the original method, while being substantially faster to run.
   
\end{abstract}

\section{Introduction}

Exponential smoothing (ETS) remains a standard forecasting procedure used in practice due to its simplicity, robustness and accuracy. In its most basic version, forecasts are produced by using the weighted sum of past observations, with the weights exponentially decaying in time. This basic version has been further extended to model trend and seasonality in either an additive or multiplicative form~\cite{holt2004Forecasting}; this is often referred to as the classical Holt-Winters method~\cite{winters1960forecasting}. Many additional extensions of the classical framework exist; perhaps most notably, \citet{gardner1985forecasting} proposed a damped version of trend to make forecasts more conservative, particularly when the forecast horizon is long. 
Modern implementations of the ETS model, such as in the R \verb|forecast| package~\cite{rForecast} and the more recent \verb|fable| package~\cite{Hyndman2021fable}, can provide practitioners with fully automatic model selection, in which no expert knowledge is required during to make forecasts. In order to facilitate the generation of probabilistic forecasts, assumptions must be made regarding the distribution of the errors, or innovations. The classical choice is to assume that the errors are normally distributed with zero mean and a constant variance over time~\cite{ForPrinPrac2021}. 

In the existing literature, most implementations of the ETS model are approached from a frequentist perspective. However, implementations within a Bayesian framework, such as those of \citet{andrawis2009new,bermudez2009multivariate,bermudez2010bayesian}, have demonstrated promising results. The drawback of Bayesian approaches has traditionally been that implementation requires a certain amount of specialised expertise, specifically with regards to posterior sampling via MCMC. The development of generic Bayesian tools such as Stan~\cite{stan2022stancore} and JAGS~\cite{plummer2003jags} has eased the pain of the modelling and programming process, but Bayesian inference still sees relatively little application within the context of exponential smoothing models. This is presumably due to the fact that existing Bayesian implementations have not shown large accuracy gains {\em vis \`{a} vis} frequentist implementations while usually being extremely slow to fit, particularly as the models become more sophisticated.

The recently proposed local and global trend (LGT) exponential smoothing model~\cite{smyl2023local} extends the classical ETS model to capture trends that grow faster than linear but slower than exponential, and relaxes the error assumption to accommodate non-normally distributed and heteroscedastic errors. This model has been able to achieve outstanding accuracy on well established benchmarks, attaining state-of-the-art performance on univariate forecasting tasks. However, while effective, a major issue with the LGT model is the high computational complexity of the Bayesian sampling procedure used to explore the posterior. In~\cite{smyl2023local} the proposed LGT models were primarily implemented via Stan~\cite{rstan2023}, with only some preliminary results for a bespoke Gibbs sampling implementation for a simplified, non-seasonal version of the model being provided. While this simplified Gibbs sampler promised speed improvements of an order of magnitude over the Stan implementations, while retaining comparable accuracy, \citet{smyl2023local} did not provide details on derivation or implementation. 

In this paper, we consolidate the seasonal and non-seasonal variants of LGT within a single model formulation, and then extend the preliminary Gibbs sampling procedure to handle this unified model. We also provide all derivations and details required for implementation. Comprehensive experiments performed on the M3 competition benchmarking dataset~\cite{makridakis2000m3} demonstrate that the proposed Gibbs sampler is not only highly accurate, but crucially, orders of magnitude faster than the original Stan implementations. This dramatic speedup has the advantage of rendering the Bayesian global and local exponential smoothing model useable in practice, yielding a procedure with acceptable time complexity that achieves state-of-the-art accuracy in many forecasting tasks. Moreover, through the novel use of the powerful horseshoe shrinkage prior for estimation of the seasonality adjustments, the resulting procedure is highly robust to the potential misspecification of seasonality. In addition to the original Stan implementation, this newly proposed Gibbs sampler is available in the R \verb|Rlgt| package on Github\footnote{https://github.com/cbergmeir/Rlgt}. This package provides a complete implementation of the proposed procedure, with detailed documentation and comprehensive examples.

This paper is structured as follows. In Section~\ref{sec:lgt}, we review the LGT model originally proposed in~\citet{smyl2023local}. In Section~\ref{sec:mcmc}, we review the basics of Bayesian inference and Monte Carlo Markov chain (MCMC) sampling approaches for posterior approximation. Section~\ref{sec:methodology} introduces the modified LGT model and details the proposed Gibbs sampling procedure for exploring the posterior distribution. A comprehensive experimental study on the M3 dataset is provided in Section~\ref{sec:experiments}. We further discuss the robustness of seasonal priors under different scenarios with an ablation study in Section~\ref{sec:ablation}. Section~\ref{sec:conclusion} concludes our work.

\section{The local and global trend model}
\label{sec:lgt}

Let $y_{t}$ denotes the realisation of the time series at time $t$. Then,  \citet{smyl2023local} define the local and global trend exponential smoothing model as:
\begin{equation}
\label{equ:lgt1} y_{t+1} \sim t(\nu,\hat{y}_{t+1}, \hat{\sigma} _{t+1}),\\
\end{equation}
where
\begin{align}    
    \label{equ:lgt2} \hat{y}_{t+1} &= l_{t}+ \gamma l_{t}^{ \rho }+ \lambda b_{t},\\
    \label{equ:lgt3} l_{t+1} &= \alpha y_{t+1}+ \left( 1- \alpha  \right) l_{t},\\
    \label{equ:lgt4} b_{t+1} &= \beta  \left( l_{t+1}-l_{t} \right) + \left( 1- \beta  \right) b_{t},\\
    \label{equ:lgt5} \hat{\sigma}_{t+1} &= \sigma \hat{y}_{t+1}^{ \tau}+ \xi.
\end{align}
$t(\nu,\mu,\sigma)$ denotes a Student $t$-distribution with degrees-of-freedom $\nu>0$, location $\mu$ and scale $\sigma>0$. Table~\ref{tab:model.params.org} details the parameters of this model and their interpretations. The one-step-ahead forecast $\hat{y}_{t+1}$ is formed as a linear combination of the (smoothed) level value, $l_t$, and local trend, $b_t$, at the previous time step. The LGT model extends the classical non-seasonal, (damped) linear trend ETS model in three major ways. First, in contrast to the classical choice of normally distributed errors, the values of series under the LGT are instead assumed to follow a Student $t$-distribution with degree-of-freedom $\nu$, location $\hat{y}_{t+1}$ and scale $\hat{\sigma}_{t+1}$. The additional degrees-of-freedom parameter $\nu$ controls the heaviness of the tail of the distribution; as $\nu$ tends to infinity, the $t$-distribution converges to a normal distribution, and as $\nu \to 0$ the tails of the $t$-distribution become increasingly heavier. Such a generalised, heavy-tailed distribution allows for the LSGT model to better capture the volatility in a time series, and provides resistance to the influence of outliers of the series. 
The second important difference from the classical ETS model is the introduction of the ``global'' trend term used when forming the one-step-ahead forecast~\eqref{equ:lgt2}. The linear weight and power parameters $\gamma$ and $\rho$ are constant over the entire series, and in this sense are global to the series. The expression $\gamma l_{t}^{ \rho }$ is a generalisation of the linear and exponential trends~\cite{smyl2023local}, and has been demonstrated to perform well in capturing trends that grows faster than linear but slower than exponential (for $\rho>0$). This term can also model the damped trend that is popular in forecasting \cite{gardner1985forecasting} if $\rho$ is taken to be negative. 
The third difference from the classical ETS models is the introduction of heteroscedasticity through the use use of a dynamic scale term $\hat{\sigma}_{t+1}$, given by~\eqref{equ:lgt5}, which is formed from a linear combination of a powered version of the prediction $\hat{y}_{t+1}$, plus an offset term $\xi>0$. In practice, the scale of errors is very likely to vary with time, and \eqref{equ:lgt5} accommodates the possibility of a larger scale of error for larger values of the series, with the rate of growth controlled by the power parameter $\tau>0$.

\begin{table}
    \centering
    \begin{tabular}{ll}
        \hline
        & Description\\
        \hline
        $\nu$ & degree-of-freedom parameter in the student $t$-distribution\\
        $\gamma$ & coefficient trend of the global trend\\
        $\rho$ & power coefficient of the global trend, in $[-0.5,1]$\\
        $\lambda$ & damping coefficient of the local trend, in $[0,1]$\\
        $\alpha$ & level smoothing parameter, in $[0,1]$\\
        $\beta$ & local trend smoothing parameter, in $[0,1]$\\
        $\zeta$ & seasonality smoothing parameter, in $[0,1]$\\
        $\sigma$ & coefficient of the size of error, positive\\
        $\tau$ & power coefficient of the size of error, in $[0,1]$\\
        $\xi$ & minimum value of the size of error, positive \\
        $b_1$ & initial local trend\\
        $s_i$ & initial seasonality, positive, i = 1,\dots, m\\
        \hline
    \end{tabular}
    \caption{Parameters of the LGT and SGT model in the original paper.}
    \label{tab:model.params.org}
\end{table}

A seasonal version of the LGT, called the seasonal global trend (SGT) model was also introduced in~\citet{smyl2023local}. Under the SGT, the time series $y_t$ is again modelled using a Student $t$-distribution, as per \eqref{equ:lgt1}, but the model forecasts $\hat{y}_{t+1}$ and $\hat{\sigma}_{t+1}$ are modified to handle multiplicative seasonality effects:
\begin{align}
    \label{equ:sgt2} \hat{y}_{t+1} &= (l_{t}+ \gamma l_{t}^{ \rho })s_{t+1}, \\
    \label{equ:sgt3} l_{t+1} &= \alpha \frac{y_{t+1}}{s_{t+1}} + \left( 1- \alpha  \right) l_{t}, \\
    \label{equ:sgt4} s_{t+m} &= \zeta  \frac{y_t}{l_t} + (1-\zeta) s_{t}, \\
    \label{equ:sgt5} \hat{\sigma}_{t+1} &= \sigma \hat{y}_{t+1}^{ \tau}+ \xi.
\end{align}
with
\begin{equation}
    \label{equ:sgt6}
    \frac{1}{m} \sum_{i=1}^m s_i = 1.
\end{equation}
The SGT model is an extension of the classical Holt-Winters model that also possesses the improvements discussed above for the non-seasonal version, and with parameters described in Table~\ref{tab:model.params.org}, and is presented in~\citet{smyl2023local} as a separate model from the non-seasonal LGT. For simplicity, the local trend component is not included in~\eqref{equ:sgt2} when forming the one-step-ahead forecasts, as empirical evidence suggested this term provided no benefit when forecasting seasonal series. The seasonality terms are multiplicative factors, and their overall effect should be not change the scale of the data; the sum constraint (\ref{equ:sgt6}) is introduced to ensure this. For further details on the LGT and SGT models we refer the reader to \citet{smyl2023local} for a thorough discussion of the model and parameter space.

While these models are flexible and powerful extensions of the classical ETS model that have demonstrated state-of-the-art performance in forecasting benchmarks, they are also substantial more complex and contain a number of additional free parameters that must be fitted. As the series to which these techniques can be applied may often be short, a Bayesian framing of the problem was used in \citet{smyl2023local} for model fitting and forecasting. Monte Carlo Markov chain (MCMC) sampling via the generic sampling tool ``Stan'' was used for posterior approximation. This process is computationally expensive, meaning that the overall fitting time, even for short series is often prohibitively long. This is the primary weakness of the LGT/SGT models in comparison to other forecasting techniques, and is partly due to the use of a generic tool which cannot directly exploit properties of the model, and partly due to the model formulation which introduces additional dependencies between model parameters, which is known to have detrimental effects on posterior exploration via MCMC. To address this weakness, this paper proposes a unified, modified LGT/SGT model and an accompanying Gibbs sampler that dramatically speeds up the MCMC sampling process. The next section discusses the fundamentals of Bayesian inference and MCMC sampling procedures, which prepares for the formal introduction of proposed sampler later.

\section{Monte Carlo Markov Chain: A Brief Review}
\label{sec:mcmc}
In the Bayesian framework, the parameters $\theta$ of the model $p(y \vbar \theta)$ are assumed to be random variables that follow a prior distribution $\pi(\theta)$. The posterior distribution of $\theta$, after seeing the sample data $y$, is given by
$$p(\theta \vbar y_1,\ldots,y_n) = \frac{p(y_1,\ldots,y_n \vbar \theta)\pi(\theta)}{p(y_1,\ldots,y_n)} \propto p(y_1,\ldots,y_n \vbar \theta)\pi(\theta),$$
where $p(y_1,\ldots,y_n \vbar \theta)$ is the likelihood function and 
\[
    p(y_1,\ldots,y_n) = \int_\Theta p(y_1,\ldots,y_n \vbar \theta) \pi (\theta) d\theta
\]
is the marginal probability of the data. From the posterior distribution, one can obtain full information about the model parameters. However, the high dimensional integral required to compute the normalising term $p(y_1,\ldots,y_n)$ is usually intractable, so in practice, a simulation approach, such as Monte Carlo Markov chain, is frequently used to approximate the posterior distribution. In the MCMC approach the posterior is approximated via a set of samples, say $\theta^{(1)}, \theta^{(2)}, \dots, \theta^{(m)}$, that are randomly drawn (simulated) from the posterior distribution. A key strength of the MCMC approach is that it is simulation consistent, in the sense that the sample distribution will converge almost surely to the exact posterior distribution as $m \to \infty$. Estimates of parameters or other posterior quantities such as intervals can readily be obtained from the collection of posterior samples. 

Recently, powerful generic Bayesian tools such as Stan have become available for Bayesian modelling and posterior exploration. These allow the non-specialist to define a Bayesian hierarchy and obtain posterior samples via Hamiltonian Monte Carlo approach. However, this generality comes at a price, as the No-U-Turn sampler (NUTS)~\cite{hoffman2014nuts} used by Stan, can be computationally expensive for even moderate numbers of model parameters. This can lead to low sampling efficiency relative to run-time, particularly if some of the model parameters exhibit a high statistical dependency. In contrast, by exploiting the specific statistical properties of a problem one can often apply computationally cheap algorithms, such as Gibbs sampling, for relatively sophisticated models. As such, depending on the structure of the model and prior distributions in question, generic Bayesian tools may be unnecessarily computationally costly, and it may be possible to obtain large computational speed-ups by developing bespoke sampling algorithms. We will now briefly review several key random number generation algorithms which are often used as building blocks for sampling algorithms.

\subsection{Gibbs sampling}

Gibbs sampling is a key MCMC algorithm. They idea underlying Gibbs sampling is that we may sample from a joint density by iteratively sampling each random variable (in our case, model parameters) iteratively from their conditional densities~\cite{robert1999monte}. This means that instead of sampling from a high-dimensional joint distribution directly, the sampling process reduces to sampling from a sequence of conditional posteriors that are potentially easier to sample from. For example, if we want to sample from a posterior distribution $p(\theta_1, \theta_2, \theta_3 \vbar y)$, then we may instead iteratively sample from $p(\theta_1 \vbar \theta_2, \theta_3, y)$, $p(\theta_2 \vbar \theta_1, \theta_3, y)$ and $p(\theta_3 \vbar \theta_1, \theta_2, y)$ (the order in which we choose to sample is irrelevant). Such a process allows us a free choice of sampling algorithm for each of the conditional distributions, and is most efficient when the conditional distributions for the parameters can be identified as some well-studied distributions, for which efficient random sampling algorithms exist. A weakness of Gibbs sampling is the high degree of dependency that is often present in the random samples, particularly if the parameters exhibit a high degree of statistical dependency. Variants of the basic Gibbs sampler have been introduced to mitigate this problem; specifically grouping or collapsing~\cite{liu1994collapsed}. When sampling the joint distribution of $\theta_1$ and $\theta_2$ is possible, we can group $\theta_1$ and $\theta_2$, and the sampling process becomes
\begin{enumerate}
    \item Sample $\theta_1$, $\theta_2$ from $p(\theta_1, \theta_2 \vbar \theta_3, y)$;
    \item Sample $\theta_3$ from $p(\theta_3 \vbar \theta_1, \theta_2, y)$.
\end{enumerate}
This will generally act to reduce correlation in the MCMC chain. Alternatively, if the marginal distribution, say $p(\theta_1, \theta_2 \vbar y) = \int p(\theta_1 \vbar \theta_2, \theta_3, y) d\theta_3$ by integrating out the auxiliary variable $\theta_3$, is easy to sample from, one can implement a collapsed Gibbs sampler:
\begin{enumerate}
    \item Sample $\theta_1$ from $p(\theta_1 \vbar \theta_2, y)$;
    \item Sample $\theta_2$ from $p(\theta_2 \vbar \theta_1, y)$;
    \item Sample the auxiliary variable $\theta_3$ from $p(\theta_3 \vbar \theta_1, \theta_2, y)$.
\end{enumerate}
In this case, by integrating out parameters we can again reduce correlation in the resulting Markov chain. In this paper, a bespoke sampler is developed based on Gibbs sampling. A majority of the conditional posteriors can be written as recognisable distributions that are straightforward to sample from using standard random sampling algorithms. For those conditional distributions for which this is not the case, we utilise either the Metropolis-Hastings algorithm or the grid sampling algorithm to generate random samples. 

\subsubsection{Scale Mixtures}
\label{sec:scale:mixtures}

As noted, Gibbs sampling is most effective when the conditional distributions can be identified as some well studied distributions. Usually, this is only the case when the prior and likelihoods are conjugate, which in turn requires that the distributions involved be members of the exponential family. In the case that non-exponential family distributions are used, it is still possible to retain conditional conjugacy by the use of continuous mixtures. The most common mixture representation is known as the scale-mixture-of-normals. A density $q(x)$ is representable by a scale-mixture-of-normals if it can be written as
\[
    q(x) = \int \phi(x \vbar m, s^2) p(s^2) d s^2,
\]
where $\phi(x \vbar m, s^2)$ denotes the probability density of a normal distribution with mean $m$ and standard deviation $s$, and $p(s^2)$ is a mixing density.
For example, the Student \textit{t}-distribution can be expressed as a normal-inverse-gamma mixture~\cite{lange1989robust}. That is, if a random variable follows a Student \textit{t}-distribution with degree-of-freedom $\nu$, location $\mu$ and scale $\tau$, i.e.,
$$y \vbar \nu, \mu, \tau \sim t(\nu, \mu, \tau),$$
then the density can be equivalently expressed as the following mixture,
\begin{align}
	\label{eq:t:smn:1}
    y \vbar \mu, \tau, \omega & \sim N(\mu, \tau^2 \omega^2), \\
	\label{eq:t:smn:2}    
    \omega^2 \vbar \nu & \sim \rm{IG} \left( \frac{\nu}{2}, \frac{\nu}{2} \right),
\end{align}
where $\rm{IG}(\alpha,\beta)$ denotes the inverse-gamma distribution with shape $\alpha$ and scale $\beta$. Here, the variable $\omega$ is often referred to as a latent or auxiliary variable. The Student \textit{t}-distribution density can be recovered by marginalising over $y$. While the introduction of a latent variable into a Bayesian hierarchy increases the number of random variables that need to be sampled, it brings the considerable advantage that the Student-$t$, which is not a member of the exponential family, is now representable as a mixture of exponential family distributions. This opens the possibility of conditional conjugacy by choice of appropriate prior distributions, which itself substantially facilitates efficient Gibbs sampling. We use this technique extensively in this paper when constructing the Gibbs sampler in Section~\ref{sec:gibbs}.

\subsection{The Metropolis-Hastings algorithm}

The Metropolis-Hastings algorithm is a versatile and powerful sampling method that is particular useful when direct sampling from the target distribution $p(\theta)$ is difficult. Given a proposal distribution, $q(\theta^* \vbar \theta)$, samples can be generated according to the following procedure~\cite{robert2016metropolishastings}:
\begin{enumerate}
    \item Generate a proposal $\theta^{*}$ from $q(\theta^{*} \vbar \theta^{\{i-1\}})$;
    \item generate $u \sim U(0,1)$, and accept $\theta^{*}$ if 
    
    $$u < \frac{p(\theta^{*})}{p( \theta^{\{i-1\}} )} \frac{q( \theta^{\{i-1\}}  \vbar \theta^{*})}{q(\theta^{*} \vbar  \theta^{\{i-1\}}  )}.$$
\end{enumerate}
Here, the quantity $\theta^{ \{i\} }$ denotes the $i$-th sample in the Markov chain. A very common choice of proposal distribution is
$$\theta^* \vbar \theta^{\{i-1\}} \sim {\rm MVN}\left( \mu \left(\theta^{\{i-1\}}\right) , \Sigma \left(\alpha, \theta^{\{i-1\}} \right)\right),$$
where ${\rm MVN}(\cdot)$ denotes a multivariate normal distribution, with $\mu(\cdot)$ and $\Sigma(\cdot)$ functions that determine the mean and covariance of the multivariate normal distribution, respectively. The parameter $\alpha$ is often called a ``step-size'', and usually controls the overall scale of the proposal; this generally needs to be chosen so that the Metropolis-Hastings procedure yields an acceptance rate around 50\% to 60\%. In this paper we use a specific variation of Metropolis-Hastings, derived from the algorithm in~\citet{titsias2018auxiliary}, in which $\mu(\cdot)$ is determined using the gradient of the negative log-likelihood and $\Sigma(\cdot)$ is determined based on both the step-size and the curvature of the prior distribution. The step-size is automatically tuned using the algorithm presented in~\citet{schmidt2020bayesian}.

\subsection{Grid sampling}
\label{sec:grid:sampling}

A finite grid approximation (``grid sampling'') is a simple and fast way to approximate a posterior distribution~\cite{mcelreath2018statistical}. Generating a sampling using a grid sampler consists of the following steps. First, generate a set of finite candidates, say $\bar{\Theta} = \{ \bar{\theta}_1, \ldots, \bar{\theta}_q \} \subset \Theta$ from the parameter space $\Theta$, and compute the corresponding posterior probability density of the conditional posterior distribution, $p(\theta \vbar \cdots)$ at each of these candidates. Then, normalise these density values and treat them as a multinomial distribution over the set $\bar{\Theta}$, i.e.,
$$\mathbb{P}(\theta = \bar{\theta}_i) \propto p(\bar{\theta}_i \vbar \cdots).$$
Finally, we draw a sample from this multinomial distribution. It is important to note that samples drawn from a grid-sampler will represent only a quantised approximation of the original continuous distribution; however, in many cases, this may be adequate if $q$ is chosen to be sufficiently large, or if model is relatively insensitive to the precise value of the parameter being sampled. An advantage of grid sampling is that each draw is independent, which helps to reduce overall correlation in the Markov chain. With an appropriately chosen grid, grid sampling can be both an efficient and accurate alternative to methods such as Metropolis-Hastings or rejection sampling.

\section{The Local-Seasonal-Global Trend (LSGT) Model}
\label{sec:methodology}

In this work, we present a unified version of the LGT and SGT, which we call the LSGT model. In addition to unify the two models into a single formulation, we also make several adjustments to the model specification; these are designed to reduce statistical dependency between model parameters, as well to simplify posterior sampling. The LSGT models observation $y_{t+1}$ as
\begin{equation}
    y_{t+1} \vbar \hat{y}_{t+1}, \hat{\sigma}_{t+1}, \nu \sim t(\nu, \hat{y}_{t+1}, \hat{\sigma}_{t+1}), \label{eq:t.dist}
\end{equation}
where
\begin{align}
    \hat{y}_{t+1} & = (l_t + \gamma l_t^\rho + \lambda b_t) s_{t+1-m},  \label{eq:lm}\\
    l_t & = \alpha \left( \frac{y_t}{s_{t-m}} \right) + (1- \alpha)l_{t-1}, \label{eq:level}\\
    b_t & = \beta (l_t - l_{t-1}) + (1 - \beta) b_{t-1}, \label{eq:local.trend}\\
    {\rm log}s_t & = \zeta {\rm log}\frac{y_t}{l_t} + (1-\zeta) {\rm log}s_{t-m}, \label{eq:seas.log} \\ 
    \hat{\sigma}_{t+1}^2 & = \chi^2 ( \phi^2 + (1-\phi)^2 l_t^{2\tau}), \label{eq:var} 
\end{align}
subject to
\begin{equation}
        \sum_{i}^m {\rm log}s_i  = 0. \label{eq:init.seasonal}
\end{equation}
\noindent The parameters of this model are described in Table~\ref{tab:model.params}. The LSGT model includes both the seasonal and non-seasonal variants with no specific distinction between the two; instead, we can recover either variant by setting some of the model parameters to specific constants. When all the seasonality factors $s_t$ are set to 1, i.e., there is no seasonal modification, the LSGT model reduces to a version of the original LGT model. Setting $\lambda = 0$ yields the seasonal version of the LSGT.

The LSGT model also makes several changes in model formulation to the LGT and SGT models of ~\citet{smyl2023local} discussed in Section~\ref{sec:lgt}. An important modification is in the way in heteroscedasticity is incorporated into the model. In the LSGT model, the conditional scale $\hat{\sigma}_{t+1}$, given by~\eqref{eq:var}, depends on the global level $l_{t+1}$ rather then the one-step-ahead forecast $\hat{y}_{t+1}$ as in the original LGT/SGT models. This has two effects: (i) it decouples the scale from the location (forecast) in the Student-$t$ distribution, reducing correlation between the parameters $(\gamma, \rho, \lambda)$ and $(\chi,\phi,\tau)$; and (ii) the conditional distribution for the weights $\lambda$ and $\gamma$ reduces to a linear regression. A second point of difference is that in the original LGT/SGT formulation, the heteroskedasticity is handled by summing the standard deviations of the homoskedastic and time varying components. This formulation is somewhat unnatural, as the variances of sums of random variables are additive, rather than their standard deviations. This formulation also introduces substantial correlation between the standard deviation of the homoskedastic component, $\xi$, and the scale $\sigma$ of the heteroskedastic component, as both must be adjusted simultaneously to maintain the same overall scale of errors. In contrast, from \eqref{eq:var} it is clear that LSGT directly models the conditional variance of $y_{t+1}$ as a scaled mixture of the homoskedastic and heteroskedastic terms. The parameter $\chi$ controls the overall scale of the error terms, while the mixing parameter $\phi$ controls how much contribution is made to the variance by the homoskedastic and heteroskedastic terms, with the model reducing to a purely heteroskedastic form when $\phi=1$. This formulation has two benefits: (i) it reduces the correlation between the parameters that determine $\hat{\sigma}_{t+1}^2$ substantially, and (ii) it allows us to easily utilise a scale-mixture representation of the Student-$t$ distribution to simplify sampling. The final modification relates to the way in which the seasonality adjustments are handled. As these quantities appear as multiplicative factors when forecasting the level (\ref{eq:level}), they are smoothed on the logarithmic scale in the LSGT model, as per~\eqref{eq:seas.log}. As the seasonal factors should not introduce an overall change in scale, the sum-constraint \eqref{eq:init.seasonal} ensures that they have a zero sum in the logarithmic scale, or equivalently, that their product is equal to one.

\begin{table}
    \centering
    \begin{tabular}{ll}
        \hline
        & Description\\
        \hline
        $\nu$ & degree-of-freedom parameter in the student $t$-distribution\\
        $\gamma$ & coefficient trend of the global trend\\
        $\rho$ & power coefficient of the global trend, in $[-0.5,1]$\\
        $\lambda$ & damping coefficient of the local trend, in $[-100,1]$\\
        $\alpha$ & level smoothing parameter, in $[0,1]$\\
        $\beta$ & local trend smoothing parameter, in $[0,1]$\\
        $\zeta$ & seasonality smoothing parameter, in $[0,1]$\\
        $\chi$ & scale of error, positive, constant for each time period\\
        $\phi$ & mixture of homoscedastic error and heteroscedastic error parameter, in $[0,1]$\\
        $\tau$ & power coefficient of the heteroscedastic error, in $[0,1]$\\
        $b_1$ & initial local trend\\
        $s_i$ & initial seasonality, positive, i = 1,\dots, m\\
        \hline
    \end{tabular}
    \caption{Parameters of the LGT model.}
    \label{tab:model.params}
\end{table}

\subsection{Prior distributions}
\label{sec:prior:distributions}

As we are using a Bayesian approach to learn the LSGT model we require the specification of suitable prior distributions over all model parameters. To avoid our choice of prior distributions introducing a strong estimation bias we choose to use weakly informative priors where appropriate. The overall error scale $\chi^2$ is assigned a standard uninformative scale-invariant prior $\chi^2 \sim (1/\chi^2) d{\chi^2}$. The coefficients $\gamma$ and $\lambda$, and the initial value of the local trend $b_1$, are all assigned weakly informative Cauchy prior distributions:
\[
	\gamma \sim C(0, s_\gamma), \; \; \lambda \sim C(0, s_\lambda) \mand b_1 \sim C(0,s_b).
\]
This choice of prior distribution {\em a priori} preferences smaller values of the coefficients, while still allowing large values to be {\em a priori} plausible. By default we take $s_\lambda=1$ and $s_\gamma = s_b = {\rm max}({\bf y})/100$, allowing the prior distributions for $\gamma$ and $b_1$ to automatically adapt to the scale of the time series.
The smoothing parameters are defined on $(0,1)$, and are assigned beta prior distributions
\[
	\alpha, \beta, \zeta \sim {\rm Beta}(a,b).
\]
The default choice of hyperparameters is $a=1$ and $b=1/2$; this distribution masses more prior probability near $\alpha=1$ (say) than $\alpha=0$. This is appropriate as small changes to $\alpha$ when $\alpha$ is close to one result in much larger changes in model response than similar magnitude changes when $\alpha$ is close to zero. The heteroscedastic mixing parameter $\phi$ is assigned a uniform distribution on $(0,1)$.

The power parameters $\tau$ and $\rho$ are sampled using a grid sampler (see Section~\ref{sec:grid:sampling}). We use a uniformly spaced grid of candidate values for both parameters (over the range of permissible values, see Table~\ref{tab:model.params}). The degrees-of-freedom parameter $\nu$ is also grid sampled; however in this case a simple uniform spacing is inappropriate. This is because the change in the behaviour of the $t$-distribution as $\nu$ varies is not uniform on the real line, i.e., increasing $\nu$ from $\nu_0$ to $\nu_0+\nu_\delta$ is not equivalent to increasing $\nu$ from $2\nu_0$ to $2\nu+\nu_\delta$, i.e., the effect of increasing $\nu$ by some amount $\delta$ depends on the value of $\nu$. Taking this into account, we choose the candidates in the $\nu$-grid so that the symmetric Kullback–Leibler (KL) divergence~\cite{kullback1951information} between all neighbouring pairs in the candidate set is equal, i.e., all neighbouring $t$-distributions are equally ``distant'' in terms of symmetric KL divergence.

Instead of using Cauchy priors as in \citet{smyl2023local}, we assign the initial seasonal factors horseshoe priors. The prior hierarchy for the horseshoe prior is
\begin{align}   
    \log s_i & \sim N(0, \psi^2_{s_i} \delta^2), \; i = 1,\dots, m, \label{eq:hs1}\\
    \psi_{s_i} & \sim C^{+}(0,1),\; i = 1,\dots, m, \label{eq:hs2}\\
    \delta & \sim C^{+}(0,1). \label{eq:hs3}
\end{align}
\noindent
An important characteristic of the horseshoe prior is its infinitely tall spike (pole) at zero. This massing of prior probability at the origin means that if the true effects are zero, or close to zero, they will be aggressively shrunk away. In the case of the log-seasonal terms, a $\log s_i=0$ implies $s_i=1$, i.e., no seasonality. This property provides the LSGT model a greater robustness to the misspecification of seasonality effects than the Cauchy priors used in the original SGT model.

\section{Posterior sampling for the LSGT model}
\label{sec:gibbs}

We now describe a Gibbs sampler for the LSGT model (\ref{eq:t.dist})--(\ref{eq:init.seasonal}) using the prior distributions discussed in Section~\ref{sec:prior:distributions}.

\subsection{Scale-mixture representations}

The continuous scale-mixture technique described in Section~\ref{sec:scale:mixtures} is employed to simplify posterior sampling. The use of scale-mixture representations  allows for conditional conjugacy even in the case of non-exponential family distributions (such as the Cauchy), at the expense of the introduction of additional latent variables that must also be sampled. We use the scale-mixture-of-normals representation of the $t$-distribution given by (\ref{eq:t:smn:1})--(\ref{eq:t:smn:2}) to rewrite the response model (\ref{eq:t.dist}) as
\begin{align*}
    y_{t+1} \vbar \hat{y}_{t+1}, \hat{\sigma}_{t+1}, \omega_{t+1}^2 & \sim N(\hat{y}_{t+1}, \hat{\sigma}_{t+1}^2 \omega_{t+1}^2), \\
    \omega_{t+1}^2 \vbar \nu & \sim \rm{IG} \left( \frac{\nu}{2}, \frac{\nu}{2} \right).
\end{align*}
Moreover, the Cauchy distribution is a special case of the Student \textit{t}-distribution with $\nu=1$. The Cauchy prior distribution for the parameter $\gamma$ with scale $s_y$ can therefore be written as
\[
    \gamma \vbar \xi_{\gamma}^2, s_{\gamma} \sim N(0, \xi_{\gamma}^2 s_{\gamma}^2) \; \; {\rm and} \; \;
    \xi_{\gamma}^2 \sim \rm{IG} \left( \frac{1}{2}, \frac{1}{2} \right),
\]
by introducing the latent variable $\xi_{\gamma}^2$, with similar representations for the parameters $\lambda$ and $b_1$. 
The half-Cauchy distribution, used in the horseshoe prior, can also be expressed as an inverse-gamma scale-mixture of inverse-gamma distributions~\cite{wand2011mean}. 
Specifically, if 
\[
y^2 \vbar \eta  \sim {\rm IG} \left( \frac{1}{2}, \frac{1}{\eta} \right) \; \; {\rm and} \; \;
\eta \sim {\rm IG} \left( \frac{1}{2}, \frac{1}{a^2} \right).
\]
then $y \vbar a \sim C^{+} (0, a)$. Using this the horseshoe prior for the seasonality factors in~\labelcref{eq:hs1,eq:hs2,eq:hs3} can be written as \cite{makalic2015simple},
\begin{align}
    {\rm log}s_i \vbar \psi^2_{s_i}, \delta^2 & \sim N(0, \psi^2_{s_i} \delta^2), \label{eq:hs.hiearchy1}\\
    \psi_{s_i}^2 \vbar \eta_{s_i} & \sim \rm{IG} \left( \frac{1}{2}, \frac{1}{\eta_{s_i}} \right), \label{eq:hs.hiearchy2}\\
    \delta \vbar \eta_{\delta} & \sim \rm{IG} \left( \frac{1}{2}, \frac{1}{\eta_{\delta}} \right), \label{eq:hs.hiearchy3}\\
    \eta_{s_1}, \dots, \eta_{s_m}, \eta_{\delta} & \sim \rm{IG} \left( \frac{1}{2}, 1 \right), \label{eq:hs.hiearchy4}
\end{align}
where $\eta_{s_1}, \dots, \eta_{s_m}$ and $\eta_{\delta}$ are latent variables. For convenience the complete Bayesian LSGT hierarchy, including the scale-mixture representations is given in Appendix \ref{appendix:hierarchy}.

\subsection{The Gibbs sampler}

Consider a time series $\mathbf{y} = (y_1, y_2, \dots, y_T)$, and the corresponding one-step-ahead forecasts produced by the LSGT model, $\hat{\mathbf{y}} = (\hat{y}_2, \dots, \hat{y}_T)$. We now present a Gibbs sampling procedure for sampling from the posterior of the LSGT model. The Gibbs sampler uses the scale-mixture-of-normals representation for the Student $t$-distribution in Steps~\ref{enum:collapse.start} to Step~\ref{enum:collapse.end}, and integrates out the latent variables, $\omega_{t}^2$, for Step~\ref{enum:no.latent} onwards. The Gibbs sampler repeatedly iterates the following steps:
\begin{enumerate}
    
    \item \label{enum:collapse.start} Sample the global variance $\chi^2$ from the inverse-gamma distribution   
    \[
        \chi^2 \vbar \cdots \sim {\rm IG}\left(\frac{T-1}{2}, \, \sum_{t=1}^{T-1} \frac{(y_{t+1}-\hat{y}_{t+1})^2}{2\omega_{t+1}^2( \phi^2 + (1-\phi)^2 l_{t}^{2\tau})} \right).
    \]
    
    Note that if the model is homoscedastic, $\phi=1$.

    \item Sample the latent variables $\omega_{t+1}^2$ from the inverse-gamma distributions
    
    \[
        \omega_{t+1}^2 \vbar \cdots \sim {\rm IG} \left(\frac{\nu+1}{2}, \, \tilde{\beta} = \frac{(y_{t+1}-\hat{y}_{t+1})^2}{2\hat{\sigma}_{t+1}^2} + \frac{\nu}{2} \right)
    \]
    
    for $t = 1,\ldots,T-1$.

    \item Sample the degrees-of-freedom $\nu$ using a grid sampler (see Section~\ref{sec:grid}).

    \item Sample the global trend coefficient $\gamma$ from the normal distribution $N(\tilde{\mu},\tilde{\sigma}^2)$, where
    
    \[
        \tilde{\mu} = \frac{1}{\tilde{\sigma}^{2}}\sum_{t=1}^{T-1} \left[ \frac{l_t^\rho s_t(y_{t+1} - (l_t + \lambda b_t) s_t)}{\omega_{t+1}^2 \hat{\sigma}_{t+1}^2} \right] \mand \tilde{\sigma}^2 = \left(\sum_{t=1}^{T-1} \frac{l_t^{2 \rho} s_t^2}{\omega_{t+1}^2 \hat{\sigma}_{t+1}^2} + \frac{1}{\xi_\gamma^2 s_\gamma^2} \right)^{-1}
    \]
    
    and $\hat{\sigma}^2_{t+1}$ is given by (\ref{eq:var}); then sample the latent variable $\xi_\gamma$ from the inverse-gamma distribution
    
    \[
        \xi_{\gamma}^2 \vbar \cdots \sim {\rm IG}\left(1, \, \frac{\gamma^2}{2 s_{\gamma}^2} + \frac{1}{2} \right).
    \]

    \item \label{enum:collapse.end} If we are using a non-seasonal model:
    
    \begin{enumerate}
        \item Sample the local trend coefficient $\lambda$ from the normal distribution $N(\tilde{\mu},\tilde{\sigma}^2)$, where
    
    \[
        \tilde{\mu} = \frac{1}{\tilde{\sigma}^{2}}\sum_{t=1}^{T-1} \left[ \frac{b_t s_t(y_{t+1} - (l_t + \gamma l_t^\rho) s_t)}{\omega_{t+1}^2  \hat{\sigma}_{t+1}^2} \right] \mand \tilde{\sigma}^2 = \left(\sum_{t=1}^{T-1} \frac{b_t^{2} s_t^2}{\omega_{t+1}^2 \hat{\sigma}_{t+1}^2} + \frac{1}{\xi_\lambda^2 s_\lambda^2} \right)^{-1},
    \]
    
    and sample the latent variable $\xi_\lambda$ from the inverse-gamma distribution
    
    \[
        \xi_{\lambda}^2 \vbar \cdots \sim {\rm IG}\left(1, \, \frac{\lambda^2}{2 s_{\lambda}^2} + \frac{1}{2} \right).
    \]

    \item Sample the initial local trend $b_1$ from the normal distribution $N(\tilde{\mu}, \tilde{\sigma}^2)$ where

    \[
        \tilde{\mu} = \frac{1}{\tilde{\sigma}^2} \sum_{t=1}^{T-1} \frac{ \lambda^2 (1-\beta)^{2(t-1)} b_1 + \lambda (1-\beta)^{t-1} (y_{t+1}-\hat{y}_{t+1})}{\omega_{t+1}^2  \hat{\sigma}_{t+1}^2} 
        \mand \tilde{\sigma}^2 = \left( \sum_{t=1}^{T-1} \frac{ \lambda^2 (1-\beta)^{2(t-1)}}{\omega_{t+1}^2 \hat{\sigma}_{t+1}^2} + \frac{1}{\xi_{b_1}^2 s_{b_1}^2} \right) ^ {-1},
    \]
    
    and sample the latent variable $\xi_{b_1}$ from the inverse-gamma distribution
    
    \[
        \xi_{b_1}^2 \vbar \cdots \sim {\rm IG}\left(1, \, \frac{b_1^2}{2 s_{b_1}^2} + \frac{1}{2} \right).
    \]

    \end{enumerate}

    \item \label{enum:no.latent} Sample the global trend smoothing parameter $\alpha$, the local trend smoothing parameter $\beta$ and (if required) the seasonal smoothing parameter $\zeta$ using a gradient-assisted Metropolis-Hastings algorithm~\cite{schmidt2020bayesian} (see Section~\ref{sec:mgrad}).

    \item \label{enum:seasonal} If we are using a seasonal model:
    
    \begin{enumerate}
     \item Sample the initial seasonal values, ${\rm log}s_i$, using a gradient-assisted Metropolis-Hastings algorithm (see Section~\ref{sec:mgrad.seasonal}).

\item Sample the horseshoe variances $\psi_{s_i}^2$ and $\delta^2$ from the inverse-gamma distributions~\cite{makalic2015simple} 
\begin{align*}
    \psi_{s_i}^2 \vbar \cdots & \sim {\rm IG} \left( 1, \frac{1}{\eta_{s_i}} + \frac{({\rm log}s_i)^2}{2\delta^2} \right), \; i = 1,\dots, m-1, \\
    \delta^2 \vbar \cdots  & \sim {\rm IG} \left( \frac{m}{2}, \frac{1}{\eta_{\delta}} + \frac{1}{2}\sum_{i=1}^{m}\frac{({\rm log}s_i)^2}{\psi_{s_i}^2} \right).
\end{align*}

    \item Sample the horseshoe latent variables $\eta_{s_i}$ and $\eta_{\delta}$ from the inverse-gamma distributions
    
\begin{align*}
    \eta_{s_i} \vbar \cdots & \sim {\rm IG} \left( 1, 1 + \frac{1}{\psi_{s_i}^2} \right),  \; i = 1,\dots, m-1, \\
    \eta_{\delta} \vbar \cdots & \sim {\rm IG} \left( 1, 1 + \frac{1}{\delta^2} \right).
\end{align*}

\end{enumerate}

\item Sample the global trend power parameter $\rho$ using a grid sampler (see Section~\ref{sec:grid}).

\item If we are using a heteroscedastic model, sample the heteroscedastic power parameter $\tau$, and the heteroscedastic mixing parameter $\phi$ using a grid sampler (see Section~\ref{sec:grid}).

\end{enumerate}
Derivations of the conditional distributions for the coefficients $\gamma$ and $\lambda$, and initial local trend $b_1$, are detailed in Appendices~\ref{sec:appendixA} and \ref{sec:b1}, respectively.

\subsubsection{Sampling $\alpha$, $\beta$ and $\zeta$}
\label{sec:mgrad}
We group $\alpha$, $\beta$ and $\zeta$ (if we are using a seasonal model) and sample them in a single step using a gradient-assisted Metropolis-Hastings algorithm~\cite{schmidt2020bayesian}. As $\alpha, \beta, \zeta \in (0,1)$, a logistic transformation is first performed to transform the parameter space into the real line, i.e., we sample $\log(\alpha/(1-\alpha))$ rather than $\alpha$. The latent variables $\omega_t^2$ are integrated out of the likelihood for better sampling convergence. 
The negative log-likelihood is
\begin{align}
\nonumber
L(\alpha, \beta, \zeta) & = \frac{\nu+1}{2} \sum_{t=1}^{T-1} {\rm log} \left(1+\frac{e_{t+1}^2}{\nu  \,  \hat{\sigma}_{t+1}^2} \right) + \frac{1}{2}\sum_{t=1}^{T-1}{\rm log} \hat{\sigma}_{t+1}^2 
\\
& = -\frac{\nu}{2}\sum_{t=1}^{T-1}{\rm log} \hat{\sigma}_{t+1}^2 + \frac{\nu+1}{2} \sum_{t=1}^{T-1} {\rm log} (\nu \, \hat{\sigma}_{t+1}^2 + e_{t+1}^2) + C, \label{eq:mh.smoothing}
\end{align}
where $e_{t+1} = y_{t+1} - \hat{y}_{t+1}$. Unlike the basic Metropolis-Hastings algorithm, the gradients of $L(\alpha, \beta, \zeta)$ with respect to $\alpha$, $\beta$ and $\zeta$ are utilized to improve the efficiency of the sampler. Note that (\ref{eq:mh.smoothing}) depends on $\alpha$, $\beta$ and $\zeta$ through the one-step-ahead predictions $\hat{y}_{t+1}(\alpha,\beta,\zeta)$ and scales $\hat{\sigma}_{t+1}(\alpha,\zeta)$. The gradients for $\alpha$ and $\beta$ for the non-seasonal model can be calculated using the chain rule, with details provided in Appendix~\ref{sec:appendixB}. For the $\zeta$ parameter, the gradients are time consuming to compute so we do not utilize them (i.e., we set the gradient for $\zeta$ to zero). As the underlying algorithm is a Metropolis-Hastings algorithm, the gradients can be computed approximately (or not computed at all, as in the case of $\zeta$) without affecting the correctness of the sampling; more accurate computations simply lead to improved efficiency.

\subsubsection{Sampling the initial seasonal factors}
\label{sec:mgrad.seasonal}
From~\eqref{eq:init.seasonal}, the last initial seasonal term $s_m$ can then be obtained by $s_m = {\rm exp} (-\sum_{i=1}^{m-1} {\rm log} s_i)$. We integrate out the latent variables, $\omega_{t}^2$, and the negative log likelihood $L(s_1,\dots, s_{m-1})$ is essentially the same as~\eqref{eq:mh.smoothing}. The relevant derivatives are presented in Appendix~\ref{sec:appendixC}.

\subsubsection{Sampling with a grid sampler}
\label{sec:grid}
A grid sampler (see Section~\ref{sec:grid:sampling}) is implemented for sampling $\nu$, $\rho$, $\phi$, and $\tau$. 
The negative log posterior for $\nu$, conditional on the latent variables $\omega_t^2$, is given by
$$L(\nu \vbar \bm{\omega}^2) = -(T-1)\frac{\nu}{2}{\rm log}\frac{\nu}{2} + (T-1){\rm log} \Gamma \left( \frac{\nu}{2} \right) + \frac{\nu+1}{2} \sum_{t=1}^{T-1} {\rm log} \omega_{t+1}^2 + \frac{\nu}{2}\sum_{t=1}^{T-1} \frac{1}{\omega_{t+1}^2},$$
where $\Gamma(\cdot)$ denotes the gamma function. We use the set of candidate $\nu$ values determined using the procedure in Section~\ref{sec:prior:distributions}.
When sampling the power parameters $\rho$ and $\tau$, and the heteroscedastic mixing parameter $\phi$, we integrate the latent variables $\omega^2_{t}$ out of the likelihood. The negative-log conditional posteriors for these parameters are given by
$$L(\rho \vbar \mathbf{\hat{y}}, \nu, \chi^2, \phi, \tau) = \frac{\nu+1}{2} \sum_{t=1}^{T-1} {\rm log} \left(1 + \frac{e_{t+1}^2}{\nu \hat{\sigma}_{t+1}^2} \right) +  {\rm log}(\rho^2+1) + C,$$
$$L(\phi \vbar \mathbf{\hat{y}}, \nu, \chi^2, \tau) = \frac{\nu+1}{2} \sum_{t=1}^{T-1} {\rm log} \left(1 + \frac{e_{t+1}^2}{\nu \hat{\sigma}_{t+1}^2} \right) + \frac{1}{2}\sum_{t=1}^{T-1}{\rm log} \hat{\sigma}_{t+1}^2 + C,$$
$$L(\tau \vbar \mathbf{\hat{y}}, \nu, \chi^2, \phi) = \frac{\nu+1}{2} \sum_{t=1}^{T-1} {\rm log} \left(1 + \frac{e_{t+1}^2}{\nu \hat{\sigma}_{t+1}^2} \right) + \frac{1}{2}\sum_{t=1}^{T-1}{\rm log} \hat{\sigma}_{t+1}^2 + C,$$
where $e_{t+1}^2 = (y_{t+1} - \hat{y}_{t+1})^2$. 
The quantities $\hat{y}_{t+1}$ and $\hat{\sigma}_{t+1}^2$ are formed using \eqref{eq:lm} and \eqref{eq:var}, respectively. The candidate values for these three parameters are set uniformly based on the corresponding parameter limits.

\section{Experiments}
\label{sec:experiments}

The proposed model extends the classical ETS model, which is a univariate forecasting procedure that does not utilize global learning across series. The M3 competition~\cite{makridakis2000m3} provides a standard benchmark dataset for univariate methods. It consists of a mix of seasonal and non-seasonal series: 645 yearly series, 756 quarterly series, 1428 monthly series, and 174 other series. We use the M3 dataset from the {\em Mcomp} R package~\cite{McompR}. Table~\ref{tab:m3data} summarizes the series lengths ($T$) and corresponding forecast horizon ($h$) in each category.

\begin{table}
    \centering
    \begin{tabular}{lrr}
        \hline
        Category & $T$ & $h$\\
        \hline
        Yearly & 14-41 & 6\\
        Monthly & 48-126 & 18 \\
        Quarterly & 16-64 & 8 \\
        Other & 63-96 & 8\\
        \hline
    \end{tabular}
    \caption{Length of observations and forecast horizon for the M3 dataset in each category.}
    \label{tab:m3data}
\end{table}

\subsection{Evaluation metrics}

Following the M3 competition, and the experimental analysis in \citet{smyl2023local}, we use the symmetric mean absolute percentage error (sMAPE) and the mean absolute scaled error (MASE) metrics to measure forecasting performance. These metrics are given by
\begin{align}
    {\rm sMAPE} & = \frac{200}{h} \sum_{t=1}^{h} \frac{|y_{T+t} - \hat{y}_{T+t}|}{|y_{T+t}| + |\hat{y}_{T+t}|} \label{eq:sMAPE},\\
    {\rm MASE} & = \frac{h^{-1}\sum_{t=1}^{h} |y_{T+t} - \hat{y}_{T+t}|}{(T-s)^{-1} \sum_{t=s+1}^T |y_t - y_{t-s}|} \label{eq:MASE},
\end{align}
respectively. The denominator in \eqref{eq:MASE} is the average error of the in-sample (seasonal) na\"ive forecasts, where $s$ denotes the periodicity; $s$ is set to one for non-seasonal series (such as yearly series), to 4 for quarterly series, and to 12 for monthly series, respectively.
Probabilistic forecasts are evaluated using the mean scaled interval score (MSIS), as per~\citet{makridakis2020m4}. The MSIS is given by
\begin{align}
    \label{eq:MSIS}
    {\rm MSIS} = \frac{
    h^{-1} \sum_{t=T+1}^{T+h} \left( q_t^{[u]} - q_t^{[l]} + \frac{2}{\alpha}(q_t^{[l]} - y_t)\mathbbm{1}_{y_t<q_t^{[l]}} + \frac{2}{\alpha} (y_t - q_t^{[u]}) \mathbbm{1}_{y_t>q_t^{[u]}} \right)
    }{
    (T-s)^{-1} \sum_{t=s+1}^T |y_t - y_{t-s}|
    },
\end{align}
where $\mathbbm{1}_x$ denotes the indicator function, which returns a one if the condition $x$ is true and a zero otherwise. The quantities $q_t^{[u]}$ and $q_t^{[l]}$ that appear in the numerator of (\ref{eq:MSIS}) are used to denote the upper and lower bounds of the prediction interval, respectively. The quantity $1-\alpha$ is the desired level of coverage; for example, $\alpha = 0.1$ if we are considering a $90\%$ prediction interval. The MSIS is an omnibus measure that penalises both the width of the forecasting interval and the attained coverage of the prediction interval.

\subsection{Results and analysis}

We consider both homoscedastic and heteroscedastic and variants of LSGT using the Gibbs sampler. The left-hand columns of Table~\ref{tab:m3-results} presents accuracy in terms of sMAPE and MASE, and the average running time per series, for both the LSGT and the original L/SGT model (with heteroscedastic errors) sampled using Stan results . The running time reported is the average running time of the models on the first 100 series in each category, executed with a single core on the same machine, for maximal comparability. The LGT Stan models have previously achieved state-of-the-art performance on the M3 dataset, as reported in \citet{smyl2023local}. Compared with the Stan sampler, the Gibbs implementations obtain slightly improved accuracy in both measures, with the improvements largest for the yearly (non-seasonal) series. When considering the model fitting time, the proposed Gibbs sampler takes significantly less computation time in comparison to the Stan implementation, and renders the LSGT model a feasible tool for deploment in practice. In regards to the different error models, the heteroscedastic models perform better than homoscedastic models on all categories except for yearly series.

\begin{sidewaystable}
\centering
\begin{tabular}{lrrr||rrrr|rr}
        \hline
         & sMAPE & MASE & Avg Runtime (s) & Below 99p & Below 95p & Below 5p & Below 1p & MSIS 90p & MSIS 98p \\
        \hline
        \multicolumn{10}{c}{Yearly series}\\
        \hline
        LSGT Gibbs (homoscedastic error) & \textbf{14.91} & 2.55 & 3.79 & 97.44 & 90.80 & 7.80 & 2.43 & 19.36 & 40.57 \\
        LSGT Gibbs (heteroscedastic error) & 14.99 & 2.50 & 4.63 & \textbf{98.94} & \textbf{94.11} & \textbf{4.96} & \textbf{1.19} & \textbf{16.47} & \textbf{27.92} \\
        LGT Stan & 15.18 & \textbf{2.48} & 60.03 & 97.16 & 91.42 & 6.23 & 2.04 & 17.38 & 32.64 \\
        \hline
        \multicolumn{10}{c}{Monthly series}\\
        \hline
        LSGT Gibbs (homoscedastic error) & 13.94 & 0.83 & 12.82 & 97.93 & 93.66 & \textbf{5.18} & 1.40 & 5.38 & 8.67\\
        LSGT Gibbs (heteroscedastic error) & \textbf{13.76} & \textbf{0.82} & 14.67 & \textbf{98.36} & \textbf{94.64} & 4.64 & \textbf{1.24} & 5.22 & 8.52 \\
        SGT Stan & 13.77 & 0.83 & 163.84 & 97.51 & 92.55 & 5.21 & 1.69 &  \textbf{5.10} &  \textbf{8.20} \\
        \hline
        \multicolumn{10}{c}{Quarterly series}\\
        \hline
        LSGT Gibbs (homoscedastic error) & 8.78 & 1.06 & 9.22 & 97.30 & 92.26 & 10.20 & 3.27 & 7.46 & 14.23 \\
        LSGT Gibbs (heteroscedastic error) &\textbf{8.78} & \textbf{1.06} & 10.70 & \textbf{97.59} & \textbf{92.97} & \textbf{8.61} & \textbf{2.12} & \textbf{7.19} & \textbf{13.06} \\
        SGT Stan & 8.87 & 1.07 & 374.12 & 96.13 & 90.16 & 11.79 & 4.76 & 7.64 & 15.95\\
        \hline
        \multicolumn{10}{c}{Other series}\\
        \hline
        LSGT Gibbs (homoscedastic error) & 4.21 & 1.70 & 5.19 & 99.64 & 97.34 & 4.02 & 0.50 & 10.8 & 17.1 \\
        LSGT Gibbs (heteroscedastic error) & \textbf{4.16} & \textbf{1.69} & 7.71 & 99.64 & \textbf{97.27} & 4.45 & \textbf{0.86} & \textbf{10.51} & \textbf{16.53} \\
        LGT Stan & 4.25 & 1.72 & 150.88 & \textbf{99.43} & 97.49 & \textbf{4.60} & 1.44 & 10.69 & 16.68 \\
        \hline
    \end{tabular}
    \caption{Accuracy results, runtime (seconds), interval coverage and MSIS scores of the models on the M3 dataset by category.}
    \label{tab:m3-results}
\end{sidewaystable}

The right-hand side of Table~\ref{tab:m3-results} provides the performance of interval coverage and the MSIS scores in terms of 90\% and 98\% prediction intervals of the two samplers. The Gibbs samplers achieves better coverage in comparison to the Stan implementation in most categories other than ``Other series''. More generally, the results suggest that the M3 series tend to be better modelled using heteroscedasticity assumptions, particularly the quarterly series. It is also worth pointing out that  \citet{smyl2023local} commented that the L/SGT models can produce slightly narrow intervals. The intervals generated by the LSGT Gibbs sampler tend to be wider as we specify a more larger space of candidate $\nu$ values in our grid. In contrast, the values used in the original Stan implementation appear to be insufficiently diverse; reducing the minimum $\nu$ in the Stan implementation could potentially fix this problem, though sampling small values of $\nu$ could also make the underlying sampling algorithm quite unstable, as Stan is known to have some issues handling heavy tailed distributions. Additionally, the implicit prior on $\nu$ used in the LSGT model (in which the candidates are equi-distant in terms of symmetric KL divergence) would be potentially quite difficult to implement in Stan, as it does not allow for easy sampling from discrete parameter spaces.  
In regards to the MSIS scores, the LSGT Gibbs sampler achieves superior results when  compared to the Stan version in all categories but monthly, and remains competitive even in this setting.

We additionally performed Wilcoxon signed rank tests of the proposed two Gibbs variants and the original Stan model. We rank the methods based on per-series performance with respect to sMAPE, MASE, MSIS90, and MSIS98. Table~\ref{tab:stats.testing} provide the average per-series ranking and the corresponding $p$-values of the testing results. From previous Table~\ref{tab:m3-results}, it can be seen that the Gibbs samplers achieved better accuracy than the original Stan L/SGT with respect to point forecast evaluation metrics. In line with the previous results, Table~\ref{tab:stats.testing} show that the Gibbs samplers rank higher than the Stan version, even though the overall performance is not statistically significant at the $0.05$ level. In terms of interval forecasting, the Stan model ranks slightly higher on average compared to both Gibbs variants. From Table~\ref{tab:m3-results}, we see that the Gibbs variants achieve higher accuracy for all but the monthly series. However, the monthly series constitute approximately half of the overall M3 dataset, and it is therefore expected that the ranking results will be largely dominated by the performance on the monthly series; additionally, rankings do not take into account the degree of difference in performance, so the larger improvements of the LSGT on yearly series, for example, are not as impactful. However, overall, the proposed Gibbs samplers are clearly highly accurate and strongly competitive with, if not superior to, the original Stan implementation in terms of forecasting metrics, while being substantially faster.

\begin{table}
    \centering
    \begin{tabular}{cccc}
    \hline
         & Gibbs (homo) - Stan & Gibbs (hetero) - Stan & Gibbs (homo) - Gibbs (hetero) \\
    \hline
    \multicolumn{4}{c}{Testing metric: sMAPE}\\
    \hline
      Method left avg rank  & 1.44 & 1.44 & 1.51\\
      Method right avg rank  & 1.56 & 1.56 & 1.49\\
      p-value & 0.75 & 0.69 & 0.94 \\
    \hline
    \multicolumn{4}{c}{Testing metric: MASE}\\
    \hline
      Method left avg rank  & 1.44 & 1.44 & 1.51 \\
      Method right avg rank  & 1.56 & 1.56 & 1.49 \\
      p-value & 0.65 & 0.62 & 0.96 \\
    \hline
    \multicolumn{4}{c}{Testing metric: MSIS90}\\
    \hline
      Method left avg rank  & 1.66 & 1.66 & 1.47 \\
      Method right avg rank  & 1.34 & 1.34 & 1.53 \\
      p-value & 0.002 & 0.003 & 0.94 \\
    \hline
    \multicolumn{4}{c}{Testing metric: MSIS98}\\
    \hline
      Method left avg rank  & 1.80 & 1.83 & 1.39\\
      Method right avg rank  & 1.20 & 1.17 & 1.61\\
      p-value & 5.93e-14 & 5.04e-16  & 0.44 \\
    \hline
    \end{tabular}
    \caption{Per-series ranking and statistical significance results in terms of the accuracy and probabilistic metrics.}
    \label{tab:stats.testing}
\end{table}

\section{Ablation study}
\label{sec:ablation}
Instead of assigning Cauchy priors to the initial seasonal factors as per the original paper LGT model in \citet{smyl2023local}, we utilise horseshoe priors (see the hierarchy ~\labelcref{eq:hs.hiearchy1,eq:hs.hiearchy2,eq:hs.hiearchy3,eq:hs.hiearchy4}). As previously discussed (Section~\ref{sec:prior:distributions}), these are a special class of priors that encourage sparsity by massing prior probability around the origin of the prior. If the log-seasonality terms are all shrunk to zero, then the multiplicative seasonality terms will be equal to one and no seasonality adjustment will occur. The motivation behind using these types of priors is to provide some robustness in the case that the user specifies seasonality, but there is no evidence in the data to support it. It is therefore of interest to test the performance of the horseshoe priors {\em vis \`{a} vis} Cauchy priors which do not encourage sparsity. 

Table~\ref{tab:ablation} summarizes the results of the ablation test. We applied the LSGT and Stan SGT/LGT models with seasonality to the monthly and yearly series. For the monthly series, we tried both horseshoe and Cauchy $C(0,1)$ priors for the LSGT with a seasonality of 12. The upper part of Table~\ref{tab:ablation} shows the results for these two priors; for this data, which likely has strong seasonal effects there is no real difference between the performance of the Cauchy and horseshoe priors, as they both have heavy tails. 
We are also interested in how robust the two priors are and their ability to distinguish if no seasonality actually occurs, even under seasonal presumptions. The lower half of Table~\ref{tab:ablation} compares the results of the non-seasonal models and seasonal models with an arbitrary periodicity of $4$ applied to the yearly series. Models that use horseshoe priors remain competitive, while models with Cauchy priors perform worse under both accuracy metrics. This suggests that the horseshoe priors are more robust and likely to achieve better results even when a seasonal model is accidentally chosen for series that may not have much evidence of seasonality. The original SGT Stan implementation used a larger scale parameter of the Cauchy prior, i.e., a heavier tail, which has poorer ability to shrink towards zero. The final entry of Table~\ref{tab:ablation} shows that this choice of prior results in very poor performance in comparison to the use of the horseshoe prior.

\begin{table}
    \centering
    \begin{tabular}{lrr}
        \hline
         & sMAPE & MASE \\
        \hline
        \multicolumn{3}{c}{Monthly series}\\
        \hline
        
        LSGT Gibbs (homoscedastic, horseshoe prior) & 13.94 & 0.83 \\
        LSGT Gibbs (heteroscedastic, horseshoe prior) & \textbf{13.76} & \textbf{0.82} \\
        
        LSGT Gibbs (homoscedastic, Cauchy prior) & 13.92 & 0.83 \\
        LSGT Gibbs (heteroscedastic, Cauchy prior) & 13.78 & 0.83 \\

        \hline
        \multicolumn{3}{c}{Yearly series}\\
        \hline
        LSGT Gibbs (homoscedastic error) & \textbf{14.91} & 2.55 \\
        LSGT Gibbs (heteroscedastic error) & 14.99 & \textbf{2.50} \\

        LSGT Gibbs (homoscedastic, horseshoe prior) & 15.35 & 2.62 \\
        LSGT Gibbs (heteroscedastic, horseshoe prior) & 15.37 & 2.55 \\

        LSGT Gibbs (homoscedastic, Cauchy prior) & 15.81 & 2.72 \\
        LSGT Gibbs (heteroscedastic, Cauchy prior) & 15.56 & 2.61 \\

        SGT Stan (Cauchy prior $C(0,4)$) & 16.56 & 2.74 \\
        
        \hline
    \end{tabular}
    \caption{Ablation study of the priors for the initial seasonal factors.}
    \label{tab:ablation}
\end{table} 

\section{Conclusion}
\label{sec:conclusion}
In this paper we have presented a fast and accurate Gibbs sampler for posterior exploration of the LSGT model. The LSGT is an extension of the classical exponential smoothing model which has the ability to capture the an heteroscedastic error structure, and super-linear but sub-exponential trends, with non-normal errors. We have combined the seasonal and non-seasonal variants presented in the work of \citet{smyl2023local} into a single formulation, and modified the model to improve statistical coherence and the efficiency of the sampling process.
In comparison to the original Stan implementation, the proposed Gibbs sampler demonstrated highly accurate performance, and importantly, is much faster, significantly reducing the computational effort required to explore the posterior distribution. The novel use of horseshoe priors in place of Cauchy priors for the seasonal factors has been demonstrated to improve the robustness of the model under both seasonal and non-seasonal conditions.

Despite the new Gibbs sampler being considerably faster than the Stan implementation, it still remains orders of magnitude slower than the classic ETS models. However, the LGT model is designed for data-scarce case, rather than the setting of big data, where global models are potentially more suitable. The promising features of the LSGT model, coupled with an efficient sampling algorithm, means that the LSGT is a feasible, and  attractive algorithm for real-world univariate, seasonal and non-seasonal, forecasting applications.

\appendix
\section{Bayesian hierarchy for the LSGT model}
\label{appendix:hierarchy}

The complete Bayesian hierarchy, including scale-mixture expansions, for the LSGT model is given below:
\begin{align*}
    y_{t+1} \vbar \hat{y}_{t+1}, \hat{\sigma}_{t+1}, \omega_{t+1}^2 & \sim N(\hat{y}_{t+1}, \hat{\sigma}_{t+1}^2 \omega_{t+1}^2), \; t = 1,\ldots,T-1\\
\end{align*}
with
\begin{align*}
    \omega_{t+1}^2 \vbar \nu & \sim \rm{IG} \left( \frac{\nu}{2}, \frac{\nu}{2} \right), \; t = 1,\ldots,T-1\\
    \nu & \sim U(\nu_l, \nu_u), \; \; (\nu_l = 1.6, \, \nu_u = 1000) \\    
    \rho & \in U(-0.5, 1), \\
    \alpha, \beta, \zeta & \sim \rm{Beta}(1, 1/2), \\
    \gamma \vbar \xi_{\gamma}^2, s_{\gamma} & \sim N(0, \xi_{\gamma}^2 s_{\gamma}^2), \; \xi_{\gamma}^2 \sim \rm{IG} \left( \frac{1}{2}, \frac{1}{2} \right), \\
    \lambda \vbar \xi_{\lambda}^2, s_{\lambda} & \sim N(0, \xi_{\lambda}^2 s_{\lambda}^2), \; \xi_{\lambda}^2 \sim \rm{IG} \left( \frac{1}{2}, \frac{1}{2} \right), \\
    b_1 \vbar \xi_{b_1}^2, s_{b_1} & \sim N(0, \xi_{b_1}^2 s_{b_1}^2), \; \xi_{b_1}^2  \sim \rm{IG} \left( \frac{1}{2}, \frac{1}{2} \right), \; \; b_1 \in (-100,1) \\
    s_{\lambda} & = 1, \; s_{\gamma}, s_{b_1} = {\rm max}({\bf y}) / 100, \\
    \chi^2 & \sim \frac{1}{\chi^2} d{\chi^2},\\
    \phi & \sim U(0, 1),\\
    \tau & \sim U(0, 1),\\
     {\rm log}s_i \vbar \psi^2_{s_i}, \delta^2 & \sim N(0, \psi^2_{s_i} \delta^2), \\
    \psi_{s_i}^2 \vbar \eta_{s_i} & \sim \rm{IG} \left( \frac{1}{2}, \frac{1}{\eta_{s_i}} \right), \\
    \delta \vbar \eta_{\delta} & \sim \rm{IG} \left( \frac{1}{2}, \frac{1}{\eta_{\delta}} \right), \\
    \eta_{s_1}, \dots, \eta_{s_m}, \eta_{\delta} & \sim \rm{IG} \left( \frac{1}{2}, 1 \right).
\end{align*}
where
\begin{align*}
    \hat{y}_{t+1} & = (l_t + \gamma l_t^\rho + \lambda b_t) s_{t+1-m}, \\
    l_t & = \alpha \left( \frac{y_t}{s_{t-m}} \right) + (1- \alpha)l_{t-1}, \\
    b_t & = \beta (l_t - l_{t-1}) + (1 - \beta) b_{t-1}, \\
    {\rm log}s_t & = \zeta {\rm log}\frac{y_t}{l_t} + (1-\zeta) {\rm log}s_{t-m}, \\ 
    \hat{\sigma}_{t+1}^2 & = \chi^2 ( \phi^2 + (1-\phi)^2 l_t^{2\tau}), 
\end{align*}
subject to
\begin{equation*}
    \sum_{i}^m {\rm log}s_i  = 0.
\end{equation*}	

\section{Derivation of normal conjugate prior}
\label{sec:appendixA}

The posterior for a normal (joint) likelihood with a normal prior also follows a normal distribution,
\begin{equation}
\label{eq:norm.format}
y_i | w \sim N((w x_i + c_i)s_i, \sigma_i^2), \; i = 1, \dots, n, \;\; {\rm and} \;\; w \sim N(0, A^2),
\end{equation}
then $w \sim N(\tilde{\mu}, \tilde{\sigma}^2)$ where
\begin{equation}
\label{eq:norm.posteior}
\tilde{\mu} = \frac{1}{\tilde{\sigma}^{2}}\sum_{i=1}^n \frac{x_i s_i(y_i - c_i s_i)}{\sigma_i^2}\;\; {\rm and} \;\; \tilde{\sigma}^2 = \left(\sum_{i=1}^n \frac{x_i^2 s_i^2}{\sigma_i^2} + \frac{1}{A^2} \right) ^ {-1}.
\end{equation}
The derivation is given as follows. The conditional posterior is obtained by multiplying the normal density,
$$p(w \vbar y_1, y_2, \dots y_n) \propto \prod_{i=1}^n \left(\frac{1}{2\pi \sigma_i^2}\right)^{1/2} {\rm exp}\left( -\frac{1}{2\sigma_i^2}(y_i-(w x_i + c_i)s_i)^2 \right) \left(\frac{1}{2\pi A^2}\right)^{1/2} {\rm exp} \left( -\frac{w^2}{2A^2} \right).$$
As only the exponential terms depend on $w$, so we drop the other terms and get
\begin{align*}
    p(w \vbar y_1, y_2, \dots y_n) & \propto \prod_{i=1}^n {\rm exp}\left( -\frac{1}{2\sigma_i^2}(y_i-(w x_i + c_i)s_i)^2 \right) {\rm exp} \left( -\frac{w^2}{2A^2} \right)\\
    & \propto {\rm exp} \left( -\frac{1}{2} \sum_{i=1}^n \frac{(y_i-(w x_i + c_i)s_i)^2}{\sigma_i^2} -\frac{w^2}{2A^2} \right).
\end{align*}
Then we expand the square term in the summation,
$$p(w \vbar y_1, y_2, \dots y_n) \propto {\rm exp} \left( -\frac{1}{2} \left(w^2 \sum_{i=1}^n \frac{x_i^2 s_i^2}{\sigma_i^2} - 2w \sum_{i=1}^n \frac{x_i s_i(y_i-c_i s_i)}{\sigma_i^2} + \frac{w^2}{A^2} + \sum_{i=1}^n \frac{(y_i-c_i s_i)^2}{\sigma_i^2} \right) \right).$$
Again, if we drop the constants,
\begin{align*}
p(w \vbar y_1, y_2, \dots y_n) & \propto {\rm exp} \left( -\frac{1}{2} \left(w^2 \sum_{i=1}^n \frac{x_i^2 s_i^2}{\sigma_i^2} - 2w \sum_{i=1}^n \frac{x_i s_i(y_i-c_i s_i)}{\sigma_i^2} + \frac{w^2}{A^2} \right) \right) \\
& \propto {\rm exp} \left( -\frac{1}{2} \left(w^2 \left(\sum_{i=1}^n \frac{x_i^2 s_i^2}{\sigma_i^2} + \frac{1}{A^2} \right) - 2w \sum_{i=1}^n \frac{x_i s_i(y_i-c_i s_i)}{\sigma_i^2} \right) \right),
\end{align*}
which is proportional to a normal distribution with mean $\tilde{\mu}$ and variance $\tilde{\sigma}^2$. If we further tidy up the above posterior,
$$p(w \vbar y_1, y_2, \dots y_n) \propto  {\rm exp} \left( -\frac{1}{2} \left(\sum_{i=1}^n \frac{x_i^2 s_i^2}{\sigma_i^2} + \frac{1}{A^2} \right) \left(w^2 - 2w \frac{\sum_{i=1}^n \frac{x_i s_i(y_i-c_i s_i)}{\sigma_i^2}}{\sum_{i=1}^n \frac{x_i^2 s_i^2}{\sigma_i^2} + \frac{1}{A^2}} \right) \right),$$
thus we get 
$$\tilde{\sigma}^2 = \left(\sum_{i=1}^n \frac{x_i^2 s_i^2}{\sigma_i^2} + \frac{1}{A^2} \right) ^ {-1},$$
and
$$\tilde{\mu} = \frac{1}{\tilde{\sigma}^{2}}\sum_{i=1}^n \frac{x_i s_i(y_i - c_i s_i)}{\sigma_i^2},$$
which matches \eqref{eq:norm.posteior}.

\section{Derivation of the conditional for $b_1$}
\label{sec:b1}

The initial value of the local trend $b_1$ is fitted for the non-seasonal model, i.e., $s_t = 1$. From \eqref{eq:local.trend}, \eqref{eq:lm} can be expressed w.r.t. $b_1$,
$$\hat{y}_{t+1} = \lambda (1-\beta)^{t-1} b_1 + c_{t+1},$$
where $c_t$ denotes the remaining constant at time $t$. The posterior distribution follows the pattern in \eqref{eq:norm.format} which can be derived by \eqref{eq:norm.posteior}, with $x_i = \lambda (1-\beta)^{t-1}$ and $s_i = 1$. However, it is much more complicated to directly calculate the remaining constant in this case. Note that $\hat{y}_i = x_i w + c_i$, the posterior mean can be expressed in an alternative form from \eqref{eq:norm.posteior} by substituting $c_i$ with $\hat{y}_i - x_i w$, so that
$$\tilde{\mu} = \frac{1}{\tilde{\sigma}^{2}} \sum_{i=1}^n \frac{x_i(y_i - c_i)}{\sigma_i^2} = \frac{1}{\tilde{\sigma}^{2}} \sum_{i=1}^n \frac{x_i^2 w + x_i(y_i-\hat{y}_i)}{\sigma_i^2},$$
given the current value of $w$.

\section{Derivation of the gradients for sampling $\alpha$ and $\beta$}
\label{sec:appendixB}
The gradients are calculated based on chain rule. We first derive the gradients for $\alpha$ as the following. With $L(\alpha, \beta, \zeta)$ defined in~\eqref{eq:mh.smoothing},
$$\dfrac{\partial L(\alpha, \beta, \zeta)}{\partial \alpha} = -\frac{\nu}{2}\sum_{t=1}^{T-1} \frac{1}{\hat{\sigma}_{t+1}^{2}} \dfrac{\partial \hat{\sigma}_{t+1}^2}{\partial \alpha} + \frac{\nu+1}{2} \sum_{t=1}^{T-1} \frac{\nu \dfrac{\partial \hat{\sigma}_{t+1}^2}{\partial \alpha} + \dfrac{\partial e_{t+1}^2}{\partial \alpha}}{\nu \hat{\sigma}_{t+1}^2 + e_{t+1}^2}.$$
Since $e_{t+1}^2 = (y_{t+1} - \hat{y}_{t+1})^2$, we get
$$\dfrac{\partial e_{t+1}^2}{\partial \alpha} = -2(y_{t+1} - \hat{y}_{t+1})\dfrac{\partial \hat{y}_{t+1}}{\partial \alpha}.$$
From~\eqref{eq:var}, we have
$$\dfrac{\partial \hat{\sigma}_{t+1}^2}{\partial \alpha} = 2\tau \chi^2 (1-\phi)^2 l_t^{2\tau-1} \dfrac{\partial l_t}{\partial \alpha}.$$
Then, we calculate $\dfrac{\partial \hat{y}_{t+1}}{\partial \alpha}$ and $\dfrac{\partial l_t}{\partial \alpha}$ recursively. According to~\eqref{eq:lm} and $s_t=1$ for the non-seasonal model, we get
$$\dfrac{\partial \hat{y}_{t+1}}{\partial \alpha} = \dfrac{\partial l_t}{\partial \alpha} + \gamma \rho l_t^{\rho-1} \dfrac{\partial l_t}{\partial \alpha} + \lambda \dfrac{\partial b_t}{\partial \alpha},$$
with \labelcref{eq:level}, we then derive
$$\dfrac{\partial l_t}{\partial \alpha} = y_t - l_{t-1} + (1-\alpha)\dfrac{\partial l_{t-1}}{\partial \alpha},$$
and
$$\dfrac{\partial b_t}{\partial \alpha} = \beta \left(\dfrac{\partial l_t}{\partial \alpha} - \dfrac{\partial l_{t-1}}{\partial \alpha}\right) + (1-\beta)\dfrac{\partial b_{t-1}}{\partial \alpha},$$
with the initial states being
$$\dfrac{\partial l_1}{\partial \alpha} = 0, \dfrac{\partial b_1}{\partial \alpha} = 0.$$

The calculation is similar for $\beta$, firstly from~\eqref{eq:mh.smoothing},
$$\dfrac{\partial L(\alpha, \beta, \zeta)}{\partial \beta} = -\frac{\nu}{2}\sum_{t=1}^{T-1} \frac{1}{\hat{\sigma}_{t+1}^{2}} \dfrac{\partial \hat{\sigma}_{t+1}^2}{\partial \beta} + \frac{\nu+1}{2} \sum_{t=1}^{T-1} \frac{\nu \dfrac{\partial \hat{\sigma}_{t+1}^2}{\partial \beta} + \dfrac{\partial e_{t+1}^2}{\partial \beta}}{\nu \hat{\sigma}_{t+1}^2 + e_{t+1}^2},$$
and
$$\dfrac{\partial e_{t+1}^2}{\partial \beta} = -2(y_{t+1} - \hat{y}_{t+1})\dfrac{\partial \hat{y}_{t+1}}{\partial \beta}.$$
The error term does not contain $\beta$, which means $\dfrac{\partial \hat{\sigma}_{t+1}^2}{\partial \beta} = 0.$ Then from~\eqref{eq:lm} we get
$$\dfrac{\partial \hat{y}_{t+1}}{\partial \beta} = \lambda \dfrac{\partial b_t}{\partial \beta},$$
as $\dfrac{\partial l_t}{\partial \beta} = 0$. Similarly with \labelcref{eq:local.trend}, we derive
$$\dfrac{\partial b_t}{\partial \beta} = (l_t - l_{t-1}) -b_{t-1} + (1-\beta)\dfrac{\partial b_{t-1}}{\partial \beta},$$
with the initial states being $\dfrac{\partial b_1}{\partial \beta} = 0.$

\section{Derivation of the gradients for sampling initial seasonal factors}
\label{sec:appendixC}
From~\eqref{eq:mh.smoothing}, we first get
$$\dfrac{\partial L(s_1,\dots, s_{m-1})}{\partial {\rm log} s_i} = -\frac{\nu}{2}\sum_{t=1}^{T-1} \frac{1}{\hat{\sigma}_{t+1}^{2}} \dfrac{\partial \hat{\sigma}_{t+1}^2}{\partial {\rm log} s_i} + \frac{\nu+1}{2} \sum_{t=1}^{T-1} \frac{\nu \dfrac{\partial \hat{\sigma}_{t+1}^2}{\partial {\rm log} s_i} + \dfrac{\partial e_{t+1}^2}{\partial {\rm log} s_i}}{\nu \hat{\sigma}_{t+1}^2 + e_{t+1}^2}.$$
Since $e_{t+1}^2 = (y_{t+1} - \hat{y}_{t+1})^2$,
$$\dfrac{\partial e_{t+1}^2}{\partial {\rm log} s_i} = -2(y_{t+1} - \hat{y}_{t+1})\dfrac{\partial \hat{y}_{t+1}}{\partial {\rm log} s_i}.$$
With~\eqref{eq:lm} we can obtain
$$\dfrac{\partial \hat{y}_{t+1}}{\partial {\rm log} s_i} = \left( \dfrac{\partial l_t}{\partial {\rm log} s_i} + \gamma \rho l_t^{\rho-1} \dfrac{\partial l_t}{\partial {\rm log} s_i} \right) s_{t+1-m} + (l_t+\gamma l_t^{\rho})\dfrac{\partial s_{t+1-m}}{\partial {\rm log} s_i},$$
where terms containing $\lambda$ are dropped for simplicity since they would equal to zero in the seasonal version. Then from~\labelcref{eq:level,eq:seas.log}, we obtain the following recursively
$$\dfrac{\partial l_t}{\partial {\rm log} s_i} = - \frac{\alpha y_t}{s_{t-m}} \dfrac{\partial {\rm log}s_{t-m}}{\partial {\rm log} s_i} + (1-\alpha)\dfrac{\partial l_{t-1}}{\partial {\rm log} s_i},$$
$$\dfrac{\partial {\rm log} s_t}{\partial {\rm log} s_i} = - \frac{\zeta}{l_t}\dfrac{\partial l_t}{\partial {\rm log} s_i} + (1-\zeta) \dfrac{\partial {\rm log} s_{t-m}}{\partial {\rm log} s_i},$$
with initial states being
$$\dfrac{\partial {\rm log}s_t}{\partial {\rm log}s_i} = 0, \;\; t \leq m, i <m, t \neq i \;\; ,\;\; \dfrac{\partial {\rm log}s_t}{\partial {\rm log}s_i} = 1, \;\; t \leq m, t=i \;\;{\rm and}\;\; \dfrac{\partial {\rm log}s_m}{\partial {\rm log}s_i} = -1,i < m;$$
and
$$\dfrac{\partial l_1}{\partial {\rm log}s_i} = \frac{y_1}{s_1}, i = 1; \;\; {\rm and}\;\; \dfrac{\partial l_1}{\partial {\rm log}s_i} = 0, 1<i<m;$$
\noindent
From~\eqref{eq:var}, we have
$$\dfrac{\partial \hat{\sigma}_{t+1}^2}{\partial {\rm log} s_i} = 2\tau \chi^2 (1-\phi)^2 l_t^{2\tau-1} \dfrac{\partial l_t}{\partial {\rm log} s_i},$$
which can be obtained based on chain rule with components already derived previously.

\bibliographystyle{plainnat}
\bibliography{references}

\end{document}